\documentclass[10pt,twocolumn,letterpaper]{article}

\usepackage{iccv}              

\usepackage{array}    
\usepackage{booktabs} 
\usepackage{multirow} 
\usepackage{xcolor}   
\usepackage{color}

\usepackage{graphicx}    
\usepackage{caption}     
\usepackage{subcaption}  
\usepackage{colortbl}
\usepackage{float}
\usepackage{algorithm}
\usepackage{amsmath}
\usepackage{adjustbox}
\usepackage{amssymb}
\usepackage{amsthm}
\usepackage{times}
\usepackage{epsfig}

\definecolor{iccvblue}{rgb}{0.21,0.49,0.74}
\usepackage[pagebackref,breaklinks,colorlinks,allcolors=iccvblue]{hyperref}

\def\paperID{5565} 
\def\confName{ICCV}
\def\confYear{2025}

\title{Boosting the Generalization and Reasoning of Vision Language Models with Curriculum Reinforcement Learning}

\author{
Huilin Deng$^{1,2}$\thanks{Work done during internship at ZTE.} \and 
Ding Zou$^{2}$\thanks{indicates equal contribution.} \and 
Rui Ma$^2$ \and 
Hongchen Luo$^3$ \and 
Yang Cao$^1$\thanks{corresponding author} \and 
Yu Kang$^1$
\\\medskip
$^1$School of Information Science and Technology, \\
University of Science and Technology of China, Hefei, Anhui, China
\\\medskip
$^2$Intelligent System Department, Zhongxing Telecom Equipment(ZTE), Changsha, Hunan, China
\\\medskip
$^3$Northeastern University, Shenyang, Liaoning, China
\\\medskip
\texttt{huilin\_deng@mail.ustc.edu.cn, zoudinghust@gmail.com, 214711069@csu.edu.cn,}\\
\texttt{luohongchen@ise.neu.edu.cn, forrest@ustc.edu.cn, kangduyu@ustc.edu.cn}
}

\begin{document}
\maketitle
\begin{abstract}
While state-of-the-art vision-language models (VLMs) have demonstrated remarkable capabilities in complex visual-text tasks, their success heavily relies on massive model scaling, limiting their practical deployment. Small-scale VLMs offer a more practical alternative but face significant challenges when trained with traditional supervised fine-tuning (SFT), particularly in two aspects: out-of-domain (OOD) generalization and reasoning abilities, which significantly lags behind the contemporary Large language models (LLMs). To address these challenges, we propose \textbf{Curriculum Reinforcement Finetuning} (\textbf{Curr-ReFT}), a novel post-training paradigm specifically designed for small-scale VLMs. Inspired by the success of reinforcement learning in LLMs, Curr-ReFT comprises two sequential stages: (1) \textbf{Curriculum Reinforcement Learning}, which ensures steady progression of model capabilities through difficulty-aware reward design, transitioning from basic visual perception to complex reasoning tasks; and (2) \textbf{Rejected Sampling-based Self-improvement}, which maintains the fundamental capabilities of VLMs through selective learning from high-quality multimodal and language examples. Extensive experiments demonstrate that models trained with Curr-ReFT paradigm achieve state-of-the-art performance across various visual tasks in both in-domain and out-of-domain settings. Moreover, our Curr-ReFT enhanced 3B model matches the performance of 32B-parameter models, demonstrating that efficient training paradigms can effectively bridge the gap between small and large models.
Our code will be released at https://github.com/ding523/Curr\_REFT.
\end{abstract}

\section{Introduction}\label{sec:intro}
\begin{figure}[t]
    \centering
    \includegraphics[width=\linewidth]{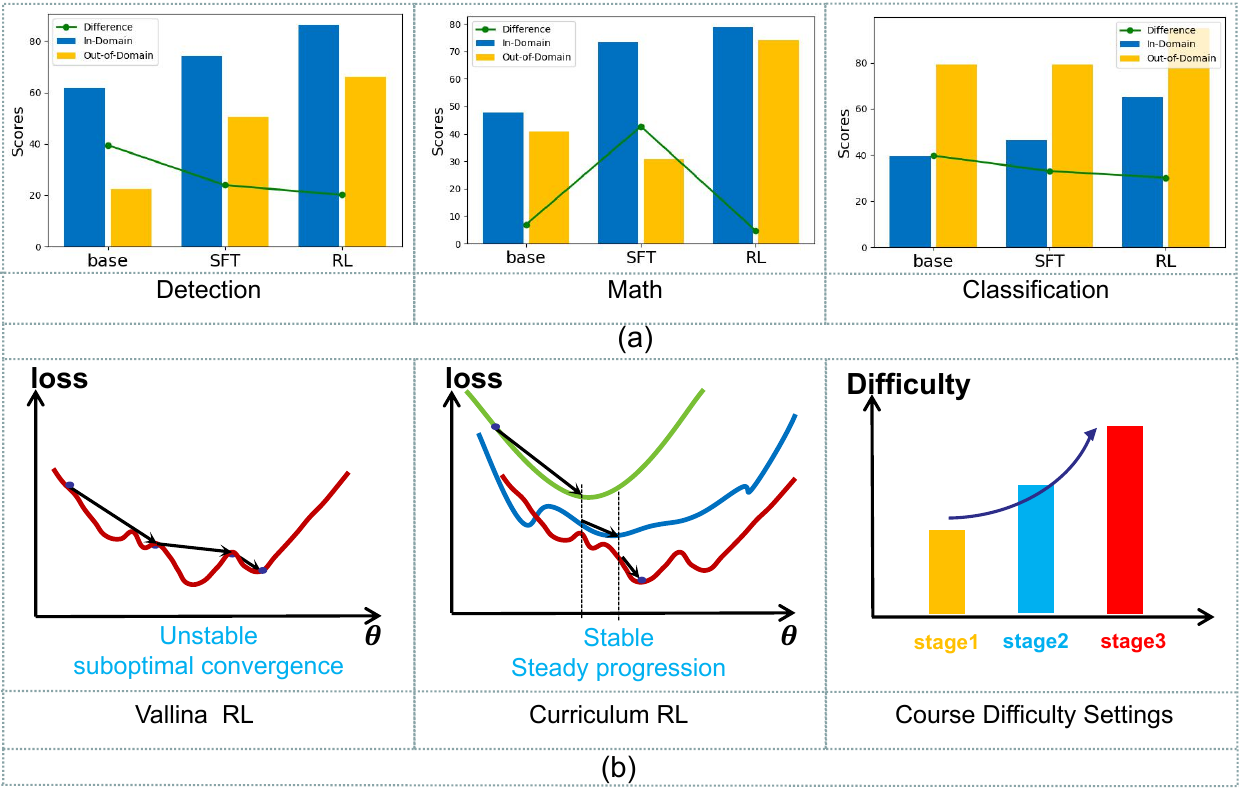}
    \vspace{-10pt}
    \caption{(a) Comparison of in-domain and out-of-domain performance between SFT and RL methods, demonstrating superior OOD generalization of RL methods across visual tasks. (b) The ``Brick Wall" phenomenon in small-scale VLMs: training instability and suboptimal convergence when facing complex examples. Our Curriculum RL ensures steady progression of model training through difficulty-aware reward design, transitioning from basic task to complex reasoning tasks.} 
    \label{fig1}
    \vspace{-10pt}
\end{figure}

Recent advances in large language models (LLMs) have catalyzed unprecedented progress in multi-modal understanding. State-of-the-art vision-language models (VLMs), exemplified by OpenAI \cite{openai_o3},\cite{openai_O1},\cite{wainwright2023instructgpt}, InterVL \cite{chen2024internvl},\cite{Self-instruct}, and QWen \cite{Qwen2VL},\cite{qwen2_5} series, have demonstrated remarkable capabilities in complex visual-text tasks. However, these achievements predominantly rely on massive model scaling ($>$32B parameters), creating substantial deployment barriers in resource-constrained environments. This limitation motivates the exploration of efficient training paradigms for small-scale VLMs (1B-7B parameters).

Current VLM training primarily utilize supervised fine-tuning (SFT) paradigms \cite{bai2022sft},\cite{ziegler2019fine} with high-quality annotated data, as exemplified by Chain of Thought (CoT) \cite{cot},\cite{wang2022self_cot}. While effective for large-scale models, SFT presents fundamental challenges for smaller architectures, manifesting in \textbf{generalization collapse} and \textbf{shallow reasoning abilities}. Specifically, task-specific SFT adaptation leads to overfitting on training set, which causes severe out-of-domain (OOD) degradation \cite{arjovsky2020ood},\cite{liu2021towards}. Moreover, complex reasoning fitting in smaller VLMs often results in superficial pattern matching rather than genuine reasoning.

The recent success of DeepSeek R1-Zero \cite{guo2025deepseek} in enhancing LLM reasoning through Group Relative Policy Optimization (GRPO) suggests a promising direction. The GRPO framework enables self-improvement through relative response comparison, naturally aligning with reasoning-intensive tasks. Given GRPO's demonstrated effectiveness in reasoning tasks, we investigate \textbf{whether RL-based post-training could enhance OOD generalization in small-scale VLMs.}

To evaluate this hypothesis, we conduct comprehensive experiments and experiments across multiple visual tasks reveal a consistent pattern: while SFT suffers from significant performance degradation on out-of-domain data, RL methods maintain robust generalization (Fig. \ref{fig1} (a), suggesting RL's potential for addressing OOD challenges.

However, our experiments also identify a critical \textbf{``Brick Wall"} phenomenon in small-scale VLMs: \textbf{models show rapid improvement on simple tasks but struggle with complex examples requiring simultaneous visual understanding and reasoning capabilities.} More concerning, when confronted with challenging cases, models experience performance degradation on previously mastered tasks. As illustrated in Fig. \ref{fig1}(b), this ``Brick Wall" effect manifests as significant training instability: the learning curve exhibits high-amplitude oscillations when encountering complex examples, ultimately leading to suboptimal convergence.

\begin{figure*}[htbp]
    \centering
    \includegraphics[width=\linewidth]{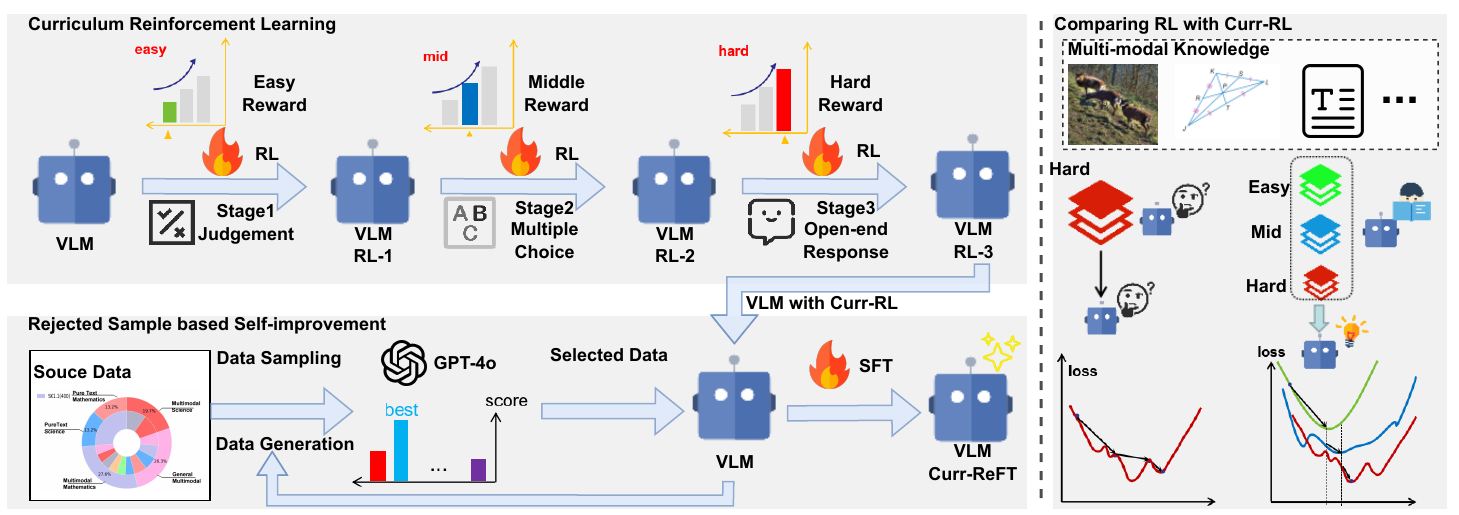}
    \vspace{-20pt}
    \caption{\textbf{Overall framework of the proposed Curr-ReFT post-training paradigm.} Curr-ReFT comprises two sequential stages: (1) Curriculum Reinforcement Learning that progressively increases task complexity with aligned reward mechanisms, and (2) Rejected Sample based Self-improvement that maintains fundamental capabilities.
    Best viewed in color. }
    \label{fig:pipeline}
    \vspace{-10pt}
\end{figure*}

To address the ``Brick Wall" phenomenon, we draw inspiration from Curriculum Learning (CL) \cite{kong2021adaptive},\cite{pentina2015curriculum},\cite{jiang2015self},\cite{huang2020curricularface}, \textbf{a training strategy that progressively exposes models to increasingly complex tasks.} We propose a novel Curriculum Reinforcement Learning paradigm that implements difficulty-calibrated rewards aligned with escalating task complexity, advancing from basic concept recognition to complex reasoning. Specifically, our Curriculum RL implements a three-stage progressive reward structure: starting with binary decision tasks (hard reward for binary yes/no responses), advancing to multiple-choice selection (intermediate reward for accurate option selection), and culminating in open-ended response (complex reward for comprehensive reasoning), as illustrated in Fig. \ref{fig:pipeline}. Moreover, to preserve fundamental language capabilities while advancing visual reasoning, we employ a rejected-sampling based self-improvement that selectively learns from both multi-modal and pure-text examples (dataset detailes in Fig. \ref{fig:data_organ} (b)). This mechanism employs reference-based response evaluation to enable targeted capability improvement while preserving existing competencies.

To this end, we propose \textbf{Curr-ReFT}, a novel post-training paradigm for enhancing reasoning and OOD generalization in small-scale VLMs. Curr-ReFT comprises two sequential stages: (1) Curriculum Reinforcement Learning that progressively increases task complexity with aligned reward mechanisms, and (2) Rejected Sample based Self-improvement that maintains fundamental capabilities through balanced learning from high equality examples.

Extensive experiments demonstrate that Curr-ReFT trained models achieves state-of-the-art performance across various visual tasks in both in-domain and out-of-domain settings and abundant public benchmarks, with our enhanced small-scale models matching the capabilities of much larger counterparts. These results provide strong evidence for the effectiveness of our Curr-ReFT paradigm.

\noindent\textbf{Our contributions} can be summarized as follows:
\begin{itemize}[leftmargin=*]
    \item \textbf{Theoretical Insight:} We demonstrate that rule-based reinforcement learning can significantly enhance the generalization capabilities of VLMs in multimodal visual perception tasks, particularly improving out-of-distribution performance without requiring additional training data.
    
    \item \textbf{Novel Framework:} We propose Curr-ReFT, a novel post-training paradigm that combines curriculum reinforcement learning with reject-sampling based self-improvement. We implement this framework in both Qwen2.5-VL-3B and Qwen2.5-VL-7B models, demonstrating its scalability and adaptability.
    
    \item \textbf{Comprehensive Evaluation:} Through extensive experiments across three datasets, we validate Curr-ReFT paradigm effective in both in-domain and out-of-domain scenarios. Our results demonstrate substantial improvements on standard benchmarks and establish new state-of-the-art performance on several key metrics.
\end{itemize}

\section{Related Work}\label{sec:related}

\subsection{Vision-Language Models}


Vision-Language Models (VLMs) have witnessed significant evolution in recent years. Early efforts focused on dual-encoder architectures, with CLIP \cite{clip}  pioneering contrastive learning for visual-textual alignment. However, they showed limitations in fine-grained visual-language alignment. The emergence of Large Language Models (LLMs) has led to a paradigm shift in VLMs. BEIT-3 \cite{Beit} introduced a unified architecture with mixture-of-experts, while LLaVA \cite{llava} advanced this trend by projecting visual features into LLMs' embedding space. Recent advances have further improved visual representation learning. Qwen-VL series \cite{bai2023qwen},\cite{Qwen2VL},\cite{qwen2_5} enhanced visual instruction tuning with multi-task training while InternVL series \cite{chen2024expanding},\cite{chen2024internvl},\cite{luo2024mono} introduced more efficient architectures. They have shown remarkable improvements in various vision-language tasks. However, optimizing for visual perception often compromises the inherent reasoning abilities of LLMs. Our ReFT addresses this challenge through multi-stage RL and rejection-sampling phase, effectively preserving the reasoning abilities with visual perception. 

\subsection{Reasoning Models}

Recent advances in reasoning models have seen significant development, especially with the integration of Monte Carlo Tree Search (MCTS) techniques \cite{browne2012survey},\cite{swiechowski2023monte} and LLMs. Innovations like Tree of Thoughts \cite{ToT} and Process-Supervised Learning \cite{lu2024autopsv} have contributed to this progress.
Building upon these foundations, OpenAI-O1 \cite{wainwright2023instructgpt} made breakthrough progress in enhancing reasoning capabilities. Following this success, numerous works \cite{openai_O1},\cite{openai_o3},\cite{brown2020language} have explored various post-training combinations of Reinforcement Learning (RL) and supervised Fine-tuning (SFT) to further improve reasoning abilities, such as Instruction-Tuned LLMs \cite{instruction-finetuned}, Self-Instruct \cite{Self-instruct}.

DeepSeek-R1-Zero \cite{guo2025deepseek} then marked a paradigm shift with its Group Relative Policy Optimization (GRPO) framework, which doesn't rely on SFT. 
GRPO uses relative performance comparisons within response groups, eliminating the need for pre-defined rules \cite{christiano2017deep} or extra critic networks \cite{schulman2017proximal}, and challenges the conventional need to combine RL with SFT \cite{gao2025comparison}. However, current reasoning models mainly focus on mathematical and coding tasks with LLMs \cite{liu2024deepseekv2},\cite{liu2024deepseekv3}, overlooking the  potential of VLMs in CV tasks. Also, most approaches are limited to large-scale architectures ($>$32B parameters), leaving the reasoning potential in smaller models unexplored.

\section{Method}\label{sec:method}



In this section, we elaborate \textbf{Curr-ReFT}, a novel post-training paradigm comprising two sequential stages: Curriculum Reinforcement Learning (Sec. \ref{method:CRL}), which orchestrates task progression through difficulty-aligned reward mechanisms, and Rejected Sample based Self-improvement (Sec. \ref{method:Rej}), which preserves fundamental capabilities via quality-guided learning. The overall framework is illustrated in Fig. \ref{fig:pipeline}.


\begin{figure}[htbp]
    \centering
    \includegraphics[width=\linewidth]{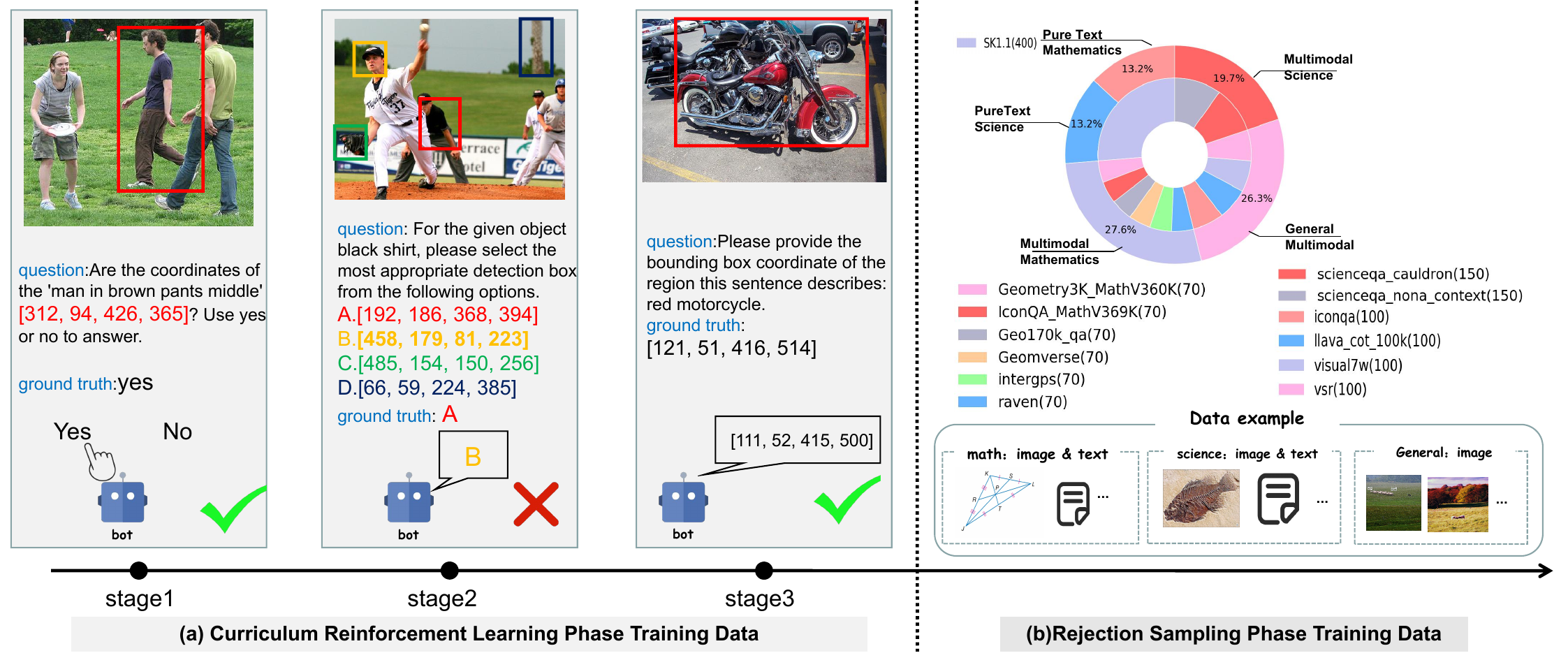}
    \caption{ \textbf{Illustration of training data organization.} (a) Examples of 3-stage progressive response formats in Curriculum Reinforcement Learning. (b) Data source in Reject-sampling SFT phase (detailed Reject-sampling pipeline in Sec. \ref{method:Rej}).}
    \label{fig:data_organ}
    \vspace{-10pt}
\end{figure}

\subsection{Preliminary}

\noindent\textbf{Reinforcement Learning with GRPO} 

Recent LLM advances have spurred interest in using reinforcement learning to boost reasoning. Reinforcement Learning from Human Feedback \cite{ouyang2022training},\cite{casper2023open} heavily depends on critic models for assessment, Reinforcement Learning with Verifiable Rewards (RLVR) hence uses direct verification functions to assess correctness. However, RLVR relies on scenario-specific rules and expert knowledge, limiting its application.

More recently, DeepSeek R1-Zero \cite{guo2025deepseek} introduces the GRPO framework, eliminating dependence on additional critic networks (PPO-based methods\cite{schulman2017proximal}). Specifically, GRPO considers the relative performance of responses rather than absolute reward values. For a given input query $q$. The framework generates $N$ distinct responses $\{o_1, o_2, ..., o_N\}$ from the current policy $\pi_{\theta}$ old and evaluates through group-wise comparison:
\begin{equation}
A_i = \frac{r_i - \text{mean}(\{r_1, \ldots, r_N\})}{\text{std}(\{r_1, \ldots, r_N\})}
\end{equation}
where $A_i$ represents the normalized relative quality of the i-th response within its group. 

\subsection{Curriculum Reinforcement Learning}  \label{method:CRL}


Curriculum Learning (CL) represents a pedagogical training strategy where models are gradually exposed to increasingly complex tasks. To address the inherent challenges of instability and convergence in reinforcement learning, we propose a novel integration of curriculum learning with GRPO that focuses on task-level progression rather than ambiguous sample-level difficulty assessment. Our key innovation lies in designing difficulty-aware reward mechanisms that align with natural task progression, advancing through three stages: \textbf{Binary Decision}, \textbf{Multiple Choice}, and \textbf{Open-ended Response} (datasets detailed in Sec. \ref{Vis_dataset}). This Curriculum Reinforcement Learning (Curr-RL) framework systematically calibrates rewards to match task complexity, enabling stable optimization across both visual perception and mathematical reasoning tasks.

\subsubsection{Stage 1: Binary Decision Learning}

In the initial stage of reinforcement learning, we adopt binary decision questions as the simplest form of task format, as shown in Fig. \ref{fig:data_organ} (a), which significantly reduces the output freedom to binary choices, making it easier to learn basic visual understanding and reasoning patterns. We explicitly instruct models to respond with only ``yes" or ``no" in our prompts. The reward function for this stage is as follows:

\begin{equation}
\mathbf{R}_\mathbf{Binary}(\mathbf{o}_{std}, \mathbf{o}_{gt}) = 
\begin{cases}
1, & \text{if } \mathbf{o}_{std} = \mathbf{o}_{gt} \\
0, & \text{otherwise}
\end{cases}
\end{equation}
where $\mathbf{o}_{std}$ represents the model's binary response and $\mathbf{o}_{gt}$ is the ground truth answer. This simple reward structure provides clear learning signals and helps establish fundamental visual-language associations.

\subsubsection{Stage 2: Multiple Choice Learning}

The second stage introduces choice questions, which require more sophisticated decision-making while maintaining structured response formats (as displayed in Fig. \ref{fig:data_organ} (a). We design different reward mechanisms for single-choice and multiple-choice scenarios to provide appropriate learning signals. For single-choice questions, we maintain a binary reward structure:

\begin{equation}
\mathbf{R}_{s}(\mathbf{o}_{std}, \mathbf{o}_{gt}) = 
\begin{cases}
1, & \mathbf{o}_{std} = \mathbf{o}_{gt} \\
0, & \text{otherwise}
\end{cases}
\end{equation}

For multiple-choice questions, we introduce a more nuanced reward function that considers partial correctness:

\begin{equation}
\mathbf{R}_{m}(\mathbf{o}_{std}, \mathbf{o}_{gt}) = 
\begin{cases}
1, & \mathbf{o}_{std} = \mathbf{o}_{gt} \\
0.2, & \mathbf{o}_{std} \subset \mathbf{o}_{gt}, |\mathbf{o}_{std}| > 0 \\
0, & \text{otherwise}
\end{cases}
\end{equation}
where $\mathbf{o}_{std}$ represents the model's selected options and $\mathbf{o}_{gt}$ is the set of correct options. This graduated reward structure encourages the model to identify correct options while maintaining the incentive for complete answers.

\subsubsection{Stage 3: Open-ended Response}


\begin{figure}[htbp]
    \centering
    \includegraphics[width=1.00\linewidth]{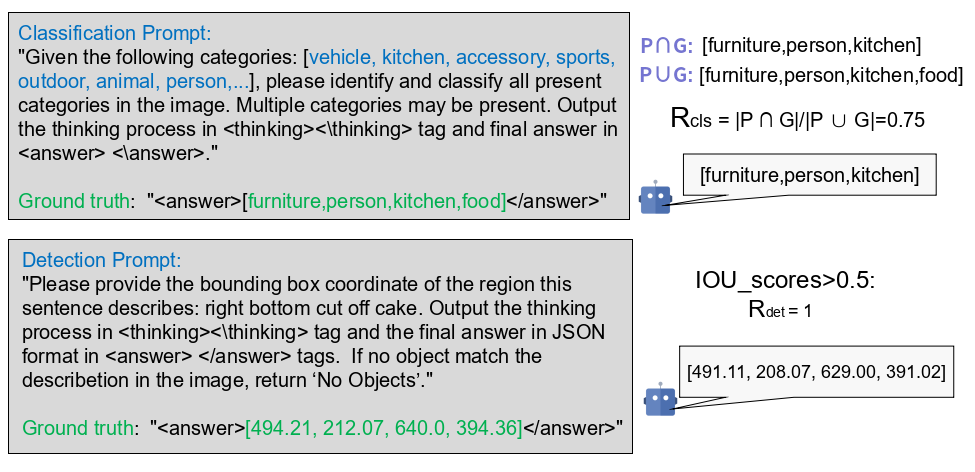}
    \caption{Verifiable Reward for visual tasks in the Open-ended Response stage. We have listed the detection and classification prompt with Verifiable Reward calculation examples.}
    \label{fig:visual_reward}
    \vspace{-10pt}
\end{figure}

Motivated by DeepSeek-R1's successful application of RL in enhancing reasoning capabilities, we extend this RL post-training to visual understanding tasks. Unlike mathematical and coding tasks with well-defined ground truth for reward computation, visual perception presents unique challenges in reward design. We develope task-specific verifiable reward functions for various visual perception tasks in the Open-ended Response stage, enabling effective RL in multi-modal contexts.

\noindent\textbf{Category Overlap Reward for Visual Classification} For classification tasks, we specifically formulate Category Overlap Reward that computes the intersection-over-union ratio between predicted and ground truth categories. This  continuous reward provides proportional credit for partial correctness, offering more informative learning signals than binary matching. Given the model's predicted categories $P = \{c_1, c_2, ..., c_m\}$ and the ground truth categories $G = \{g_1, g_2, ..., g_n\}$, where $c_i$ and $g_j$ represent individual category labels. The IoU-based classification reward is calculated based on their intersection and union:

\begin{equation}
\mathbf{R}_{acc\_cls} =  \frac{|P \mathbf{\cap} G|}{|P \mathbf{\cup} G|}=\frac{|\{c_i | c_i \in P \text{ and } c_i \in G\}|}{|\{c_1,...,c_m\} \mathbf{\cup} \{g_1,...,g_n\}|},
\end{equation}
where $|P\mathbf{\cap}G|$ represents the number of correctly predicted categories, and $|P\mathbf{\cup}G|$ represents the total number of unique categories in both sets combined. This reward mechanism provides a continuous value in [0,1], better reflecting partial correctness in multi-label scenarios compared to binary rewards. The classification reward $R_{cls}$ combines accuracy and format compliance.

\noindent\textbf{IOU rewards for Visual Detection} For object detection tasks, we design a comprehensive reward function that evaluates both localization accuracy. The reward mechanism considers three key aspects: spatial accuracy, prediction reliability, and response format compliance.

Given a set of predicted bounding boxes $B_{student} = \{b_1, b_2, ..., b_n\}$ with corresponding confidence scores $f = \{f_1, f_2, ..., f_n\}$, and ground truth boxes $B_{gt} = \{b^{gt}_1, b^{gt}_2, ..., b^{gt}_m\}$, we first establish box-level correspondences through IoU matching. By applying a threshold $\tau$, we filter out low-quality matches where $iou_i < \tau$. The localization accuracy reward $R_{loc}$ is then computed as the mean IoU of the remaining valid matches:

\begin{equation}
\mathbf{R}_{Iou} = \frac{1}{|\mathcal{V}|}\sum_{i \in \mathcal{V}} iou_i, \quad \mathcal{V} = \{i | iou_i \geq \tau\}
\end{equation}
where $\mathcal{V}$ denotes the set of valid matches and $|\mathcal{V}|$ represents the number of valid matches. To encourage accurate object localization, we further discretize the IoU-based reward using a threshold of 0.5:

\begin{equation}
\mathbf{R}_{acc\_det} = 
\begin{cases}
1, & \text{if } \mathbf{R}_{Iou} > 0.5 \\
0, & \text{otherwise}
\end{cases}
\end{equation}

The final detection reward $\mathbf{R}_{det}$ combines both localization accuracy and format compliance:

\begin{equation}
\mathbf{R}_{det} = \mathbf{R}_{acc\_det} + \mathbf{R}_{format}
\end{equation}
where $\mathbf{R}_{acc\_det}$ evaluates spatial localization accuracy and $R_{format}$ verifies response format compliance.

\subsection{Rejected Sample based Self-improvement}\label{method:Rej}
To preserve model competencies while maintaining reasoning capabilities, we propose a Rejected Sample-based Self-improvement methodology grounded in curriculum reinforcement learning principles. This approach comprises two essential components: High-Quality Data Sampling and Self-Improvement Training, enabling systematic enhancement while preserving fundamental model capabilities.

\subsubsection{High-Quality Data Sampling}

The data preparation process involves systematic sampling from a comprehensive dataset. Utilizing GPT-4-O as the reward model, we evaluate generated responses against multiple criteria: accuracy, logical consistency, format compliance, and linguistic fluency. Responses are quantitatively assessed on a 0-100 scale, with those surpassing a threshold of 85 being integrated into the enhanced dataset alongside their corresponding queries. The resultant curated dataset encompasses 1,520 high-quality examples across diverse domains: pure text mathematics, science, multimodal mathematics, and general knowledge (Fig. \ref{fig:data_organ}).

\subsubsection{Self-Improvement Training}

Based on the curated dataset, we implement self-improvement optimization to further enhance fundamental, while preserving reasoning capabilities. The optimization objective is formulated as:
\begin{equation}
L_{\text{SFT}}(\theta) = -\frac{1}{N} \sum_{i=1}^{N} \sum_{pos=0}^{SEQ} \sum_{j=1}^{C} y_{ij}^{pos} \log P(y_{j}^{pos} \mid x_i; \theta)
\end{equation}

Where:
\begin{itemize}
\item $\theta$ represents the model parameters.
\item $N$ denotes the number of samples.
\item $SEQ$ represents the length of tokens for the question's answer
\item $C$ signifies the length of the vocabulary.
\item $y_{ij}^{pos}$ indicates the label generated by the model itself at the position pos.
\item $P(y_{j}^{pos} \mid x_i; \theta)$ represents the predicted probability of class j at position pos for sample i.
\end{itemize}

\section{Experiments}\label{sec:exp}

Aiming to answer the following questions, we conduct extensive experiments and test on abundant benchmarks:
\begin{itemize}
    \item \textbf{RQ1:} How does RL perform comparing to traditional SFT in traditional CV tasks?
    \item \textbf{RQ2:} How does models trained with Curr-Reft perform compared to current mainstream VLMs?
    \item \textbf{RQ3:} How do the main components in Curr-ReFT affect its effectiveness?
    \item \textbf{RQ4:} Is the Curr-ReFT still effective when scaling the parameters?
\end{itemize}

\subsection{Experiment Settings}

\subsubsection{Datasets and Metrics} \label{Vis_dataset}
To evaluate the effectiveness and generalization of RL on VLMs, we constructed a comprehensive evaluation framework across three distinct multimodal tasks:

\noindent\textbf{Visual Detection:} we sampled 3,000 training and 1,000 in-domain testing images from RefCOCO \cite{Refcoco}, with additional 1,000 out-domain samples from Refgta \cite{Refgta} for evaluating object localization.

\noindent\textbf{Visual Classification:} the dataset comprises 3,000 training images from RefCOCO \cite{Refcoco} and RefCOCOg \cite{Refcocog}, with 1,000 in-domain and 1,000 out-domain (Pascal-VOC \cite{pascal_voc}) testing samples for evaluating visual categorization ability.

\noindent\textbf{Multimodal Mathematical Reasoning:} 
the multi-modal math dataset covers geometry proofs and visual math problems, which has 3,000 training and 1,000 testing samples from Math360K \cite{shi2024math360k} and Geo170K \cite{gao2023geo170k}, plus 500 CLEVER-70k-Counting samples for out-domain testing.



We use accuracy as unified evaluation metric, defined as correct predictions over total test samples. For detection, a prediction is correct if the IoU between predicted and ground truth boxes exceeds 0.5. In classification, predictions matching ground truth labels are considered correct.

\subsubsection{Benchmarks}
To give a reasonable result, we evaluate our trained models on the following authoritative benchmarks:
\begin{itemize}
    \item \textbf{MathVisa \cite{lu2023mathvista}} is a comprehensive mathematical benchmark containing 6,141 examples across 31 datasets.
    \item \textbf{MATH \cite{MATH}} comprises 12,000 high school competition-level problems, which spans over arithmetic, geometry, number theory, probability and statistics.
    \item \textbf{AI2D\cite{hiippala2021ai2d}} contains over 5000 grade school science diagrams with over 150000 rich annotations and more than 15000 corresponding multiple choice questions.
    \item \textbf{MMVet \cite{yu2023mm}} focuses on complex reasoning through 6 assessments(OCR, visual grounding, commonsense reasoning, visual recognition, inference and spatial understanding),utilizing an LLM-based evaluator for unified scoring.
    \item \textbf{MMBench \cite{liu2024mmbench}} specializes in fundamental multimodal abilities through 3,000 multiple-choice questions, evaluating perception-based tasks (object detection, attribute recognition, spatial relationships).
    \item \textbf{OCRBench \cite{liu2024ocrbench}} converts text in images into readable format, is a fundamental task in document understanding.
    \item \textbf{LLaVABench \cite{llava}} focus on the evaluation of generalizability to novel domains. It consists of a diverse set of 24K images with 60K questions in total. 
    
\end{itemize}

\subsubsection{Baselines}

To comprehensively evaluate our post-train approach, we conduct extensive experiments against state-of-the-art vision-language models (VLMs) across different parameter scales (3B to 32B). The baseline models include: 1) Small-scale models (3B-4B): Qwen2.5-VL-3B and InternVL2\_5-4B. 2) Medium-scale models (7B-8B): Qwen2-VL-7B, Qwen2.5-VL-7B, and InternVL2\_5-8B
3) Large-scale models ($>$20B): InterVL2-26B and LLAVA-next-qwen-32b. Specifically, we evaluate all models across visual understanding tasks and mathematical reasoning in mix datdasets and comprehensive benchmarks to ensure a comprehensive comparison.

\subsubsection{Implement Details}

All experiments are conducted on NVIDIA A800 GPUs. The majority of experiments use Qwen2.5-VL-3B as the base model, trained on a single server with 8 A800 GPUs using batch size of 8. For scaling experiments, we employed Qwen2.5-VL-7B as the base model, utilizing two servers with 8 A800 GPUs each. The hyperparameters are set as follows: (1) Learning rates: 2e-5 for RL (GRPO) training, 2e-7 for rejection sampling phase, and 1e-6 for baseline SFT experiments. (2) Maximum pixel size: 401,408. (3) GRPO training steps: 2,500. Specially, we use Qwen2.5-VL-3B as the reference model in standard experiments and Qwen2.5-VL-7B for scaling experiments.

\subsection{Generalization Verification of RL (RQ 1)}


\begin{table}[htbp]
\centering
\small  
\begin{tabular}{l|ccc|ccc}
\hline
\rule{0pt}{2.5ex} 
\multirow{2}{*}{\scriptsize Method} & \multicolumn{3}{c|}{\cellcolor{gray!15}In-domain} & \multicolumn{3}{c}{\cellcolor{yellow!15}Out-domain} \\
\cline{2-7}
\rule{0pt}{2.5ex} 
& \cellcolor{gray!15}Det & \cellcolor{gray!15}Math & \cellcolor{gray!15}Cls & \cellcolor{yellow!15}Det & \cellcolor{yellow!15}Math & \cellcolor{yellow!15}Cls \\
\hline
Base  & \cellcolor{gray!15}61.8 & \cellcolor{gray!15}47.8 & \cellcolor{gray!15}39.6 & \cellcolor{yellow!15}22.3 & \cellcolor{yellow!15}40.8 & \cellcolor{yellow!15}79.3 \\
+SFT & \cellcolor{gray!15}75.2 & \cellcolor{gray!15}73.5 & \cellcolor{gray!15}50.2 & \cellcolor{yellow!15}52.3 & \cellcolor{yellow!15}30.8 & \cellcolor{yellow!15}77.2 \\
+RL & \cellcolor{gray!15}88.3 & \cellcolor{gray!15}78.8 & \cellcolor{gray!15}62.9 & \cellcolor{yellow!15}64.2 & \cellcolor{yellow!15}74.1 & \cellcolor{yellow!15}94.7 \\
+Curr-RL & \cellcolor{gray!15}\textbf{90.6} & \cellcolor{gray!15}\textbf{82.8} & \cellcolor{gray!15}\textbf{66.8} & \cellcolor{yellow!15}\textbf{67.1} & \cellcolor{yellow!15}\textbf{78.4} & \cellcolor{yellow!15}\textbf{96.6} \\[0.5ex] 
\hline
\end{tabular}
\caption{Performance Comparison: In/Out-domain Performance (\%). Base model choose the Qwen2.5-VL-3B. Notably, `Det' and `Cls' denote detection and classification, respectively.}
\label{tab:comprehensive}
\end{table}

\begin{figure}[htbp]
\centering

\begin{subfigure}[b]{0.48\linewidth}
    \centering
    \includegraphics[width=\textwidth]{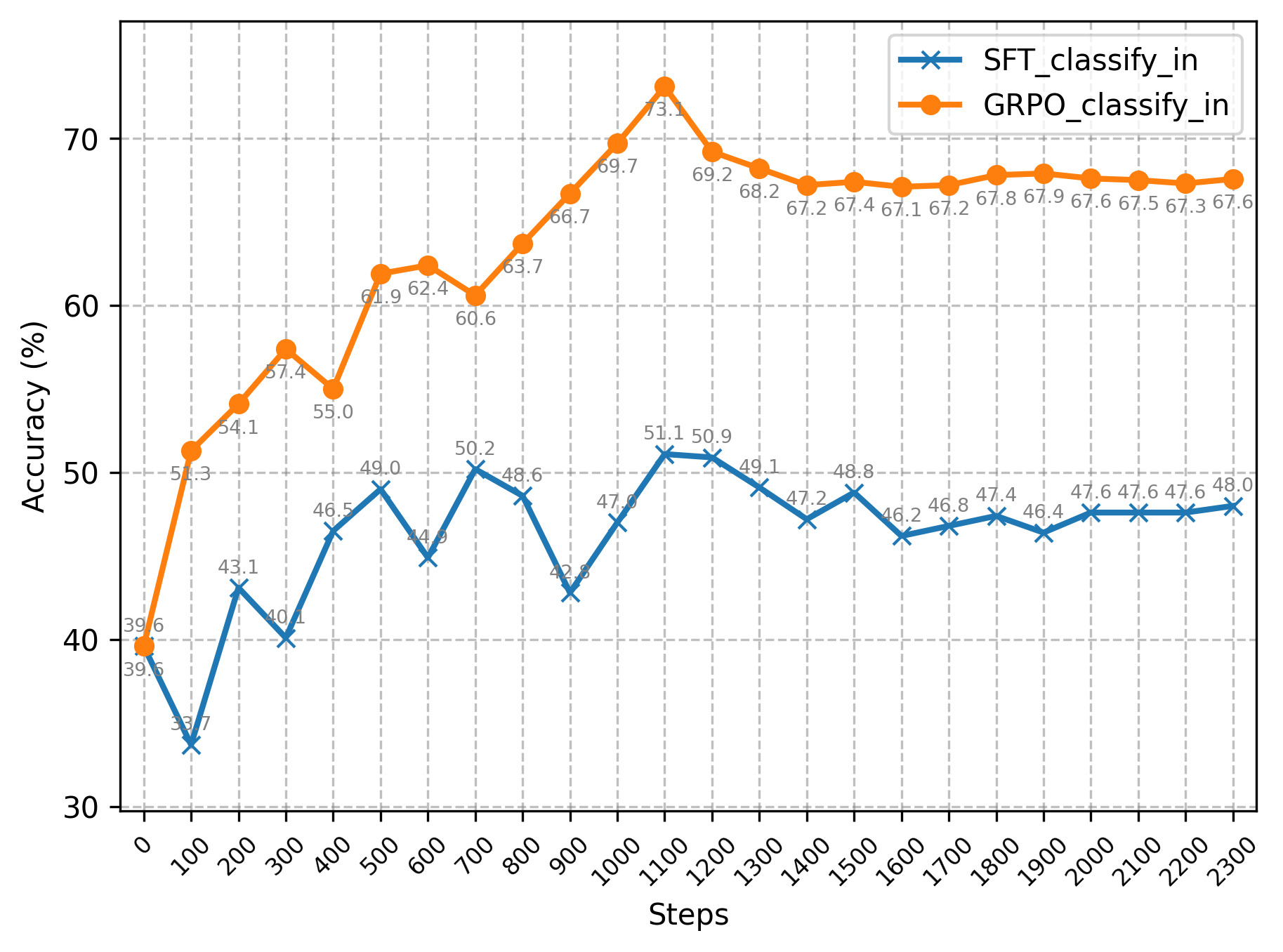}
    \caption{In-domain Classification}
    \label{fig:classify_in}
\end{subfigure}
\hfill  
\begin{subfigure}[b]{0.48\linewidth}
    \centering
    \includegraphics[width=\textwidth]{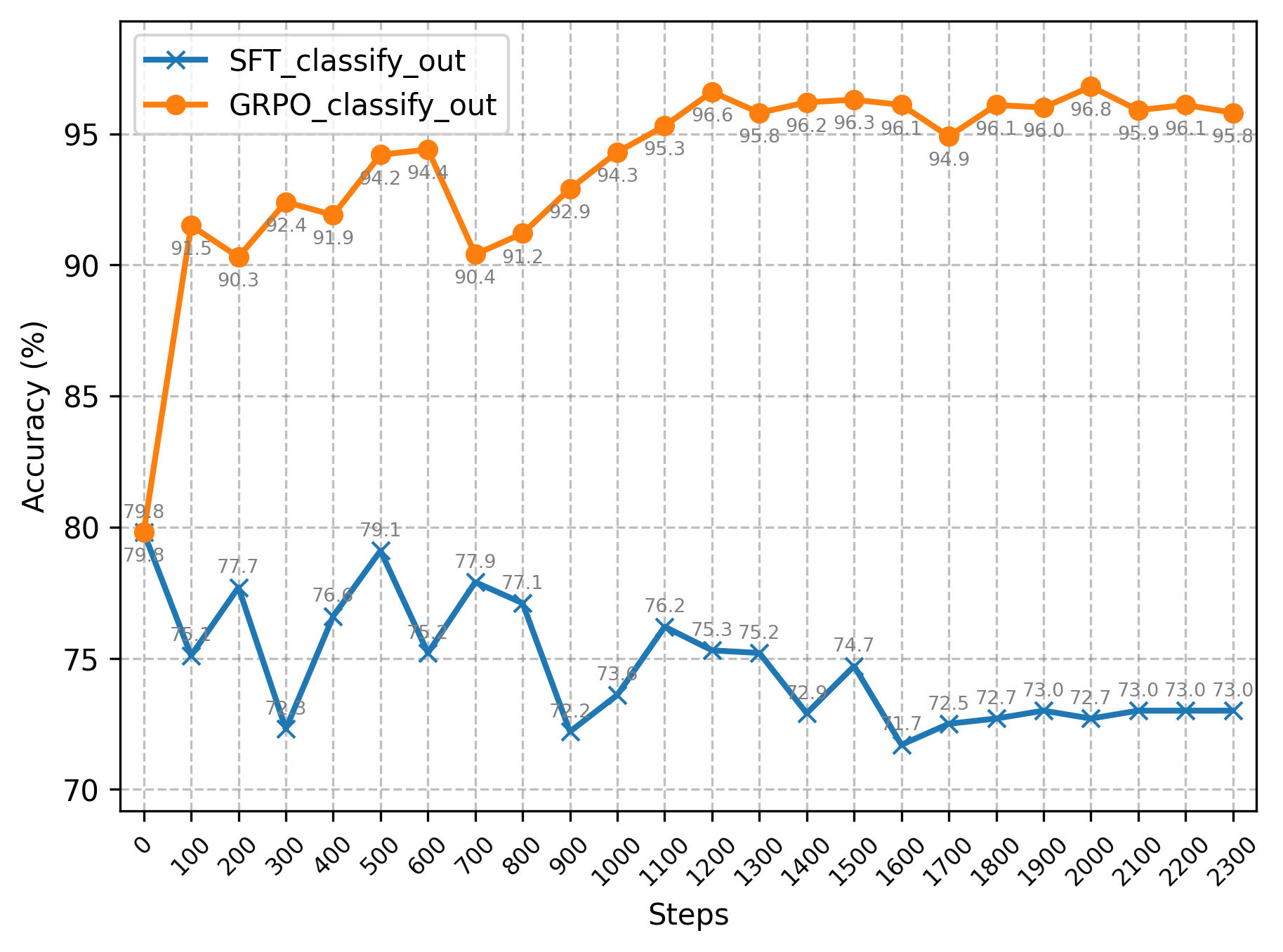}
    \caption{Out-domain Classification}
    \label{fig:classify_out}
\end{subfigure}

\vspace{0.2cm}  

\begin{subfigure}[b]{0.48\linewidth}
    \centering
    \includegraphics[width=\textwidth]{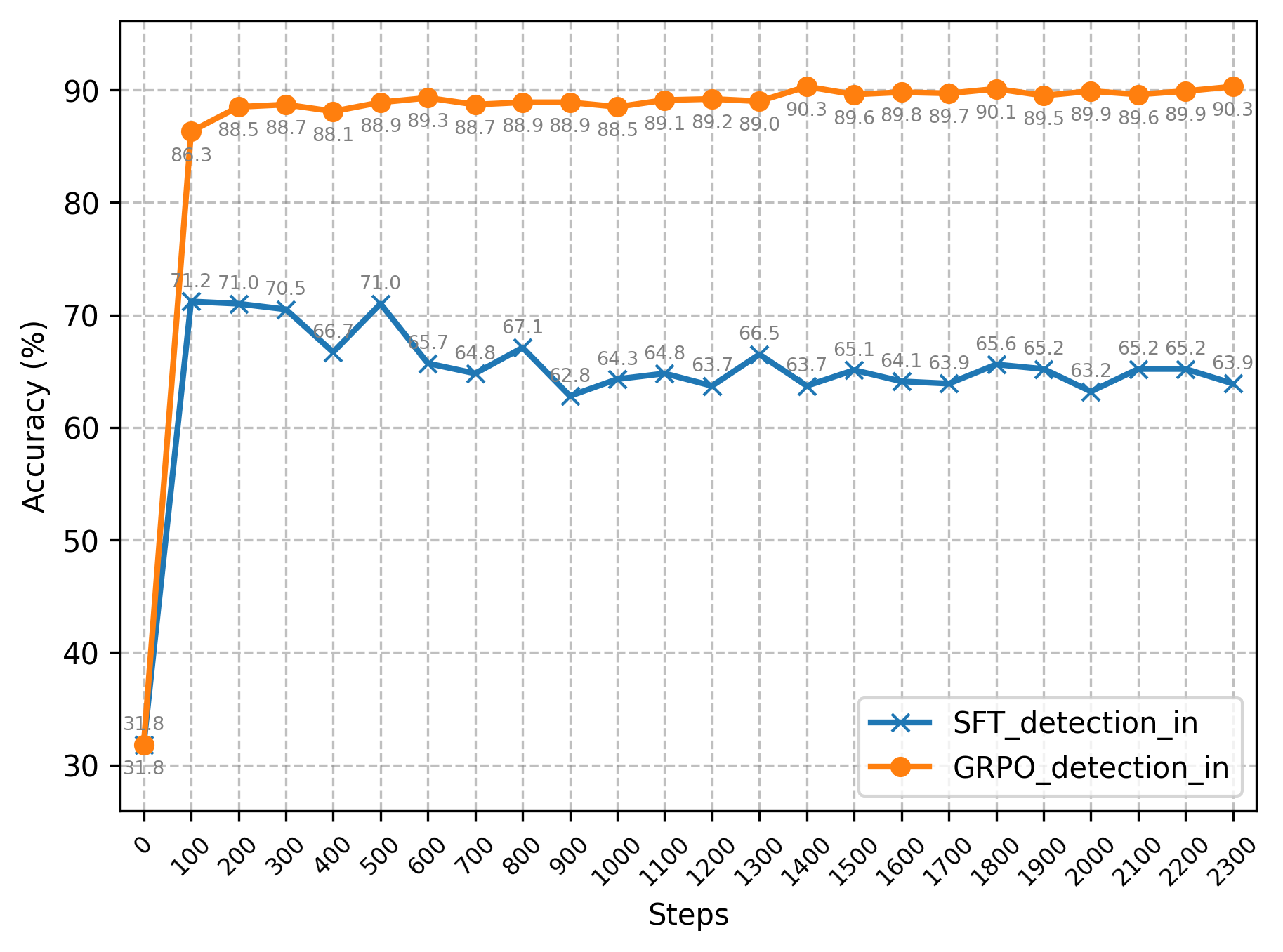}
    \caption{In-domain Detection}
    \label{fig:detection_in}
\end{subfigure}
\hfill  
\begin{subfigure}[b]{0.48\linewidth}
    \centering
    \includegraphics[width=\textwidth]{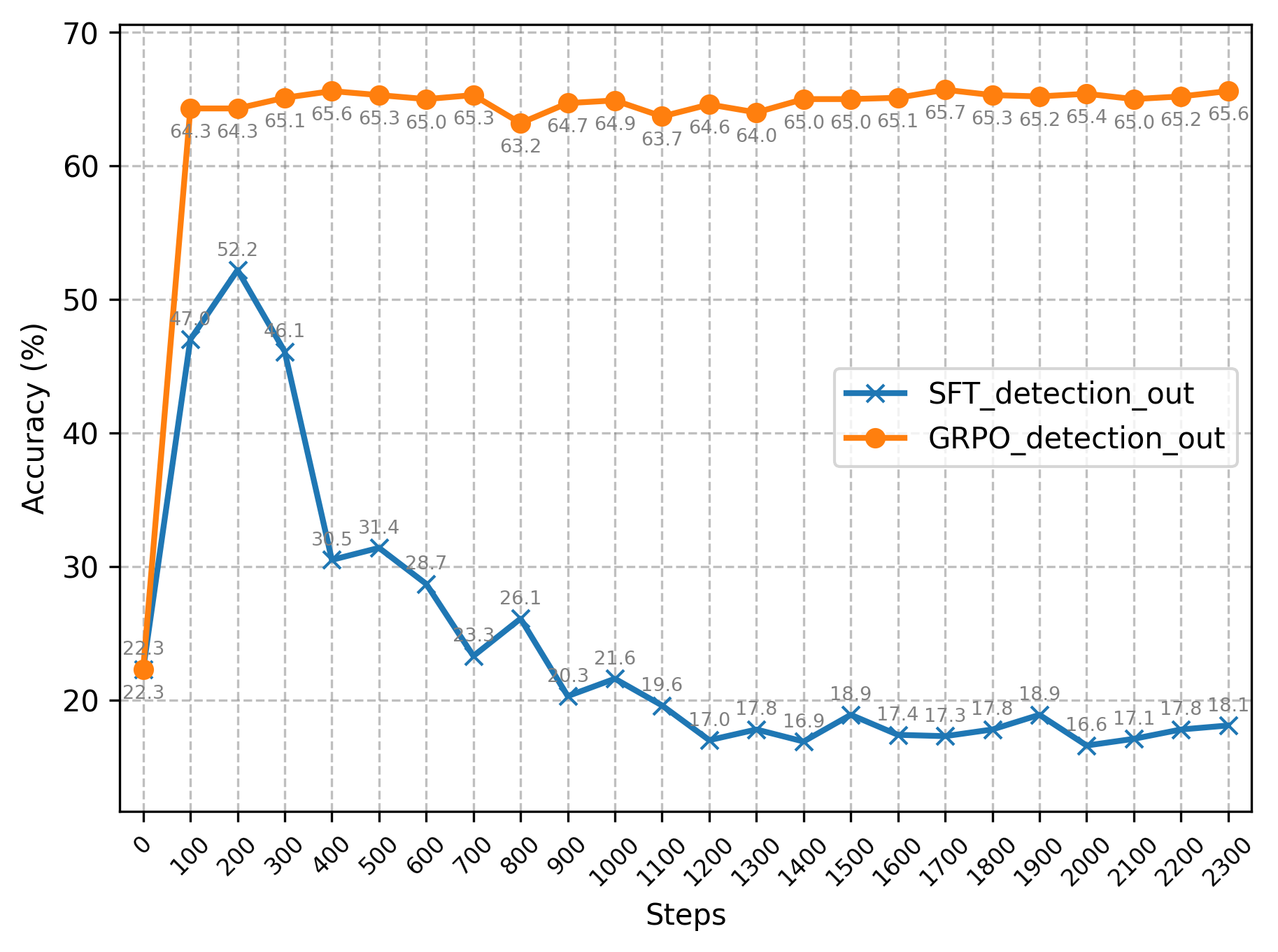}
    \caption{Out-domain Detection}
    \label{fig:detection_out}
\end{subfigure}
\caption{Empirical evaluation of SFT versus GRPO across in-domain and out-of-domain tasks.}
\label{fig_sft_vs_grpo}
\end{figure}

\begin{figure}[htbp]
    \centering
    \includegraphics[width=\linewidth]{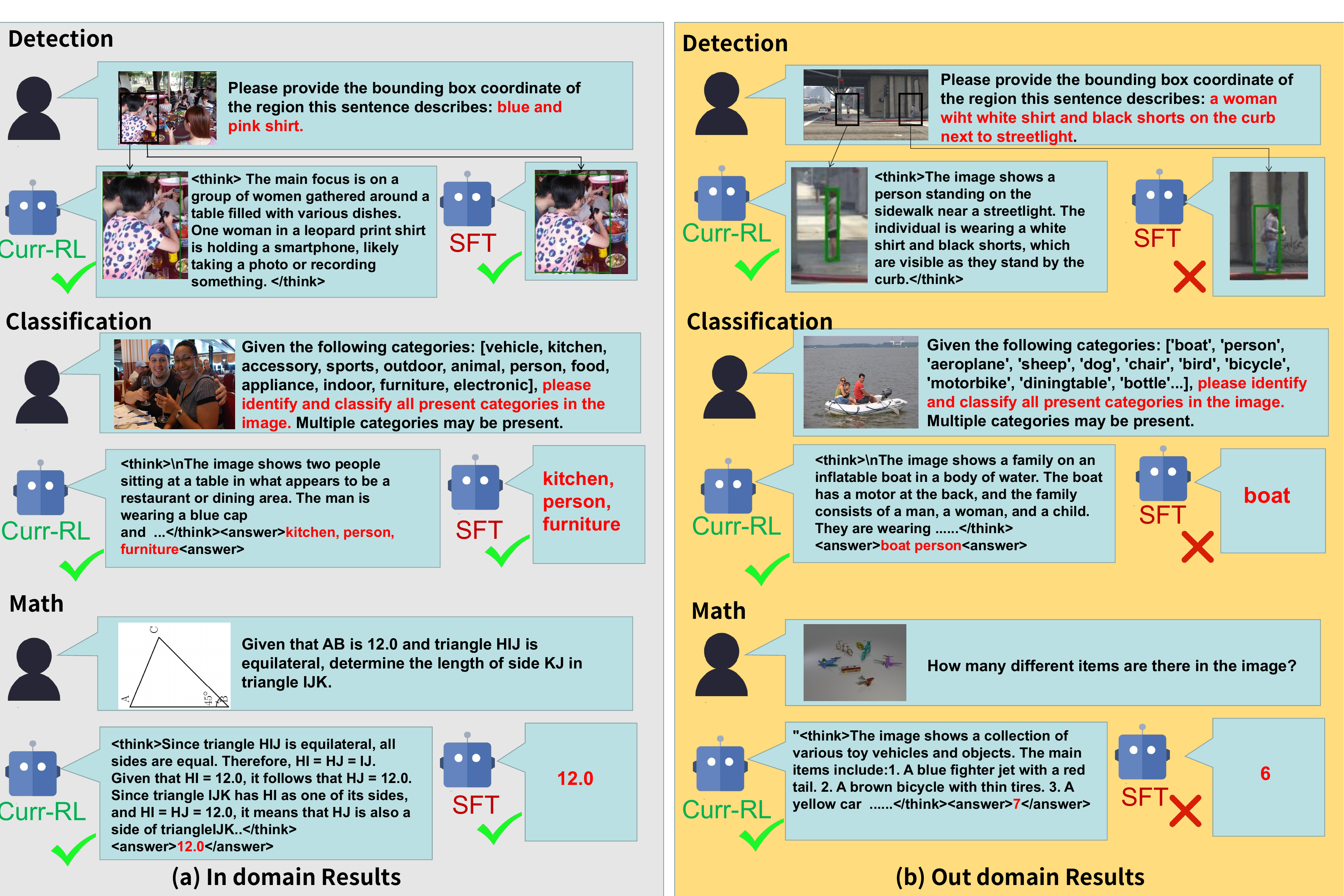}
    \caption{Qualitative comparison between our method and SFT baseline. Thinking significantly improves reasoning ability.}
    \label{compare_viosn}
\end{figure}

\begin{table}[th]
\setlength{\tabcolsep}{2pt}  
\centering
\begin{tabular}{>{\raggedright\arraybackslash}p{2.4cm} >{\centering\arraybackslash}p{0.8cm} >{\centering\arraybackslash}p{0.8cm} >{\centering\arraybackslash}p{0.9cm} >{\centering\arraybackslash}p{0.9cm} >{\centering\arraybackslash}p{0.9cm} >{\centering\arraybackslash}p{0.9cm}}
\toprule
\multirow{2}{*}{\textbf{Methods}} & \multicolumn{2}{c}{\textbf{Math}} & \multicolumn{2}{c}{\textbf{Detection}} & \multicolumn{2}{c}{\textbf{Classification}} \\
\cmidrule(lr){2-3} \cmidrule(lr){4-5} \cmidrule(lr){6-7}
& \textbf{In} & \textbf{Out} & \textbf{In} & \textbf{Out} & \textbf{In} & \textbf{Out} \\
\midrule
\multicolumn{7}{l}{\textbf{Base Models}} \\
\midrule
Qwen2.5-VL-3B & 71.3 & 17.8 & 31.8 & 22.3 & 39.6 & 79.8 \\
InternVL2\_5-4B & 58.4 & 30.3 & 31.2 & 24.5 & 21.5 & 78.9 \\
Qwen2.5-VL-7B & 75.9 & 34.9 & 41.3 & 27.8 & 49.7 & 86.2 \\
Qwen2-VL-7B & 74.1 & 26.7 & 40.9 & 43.6 & 28.5 & 61.3 \\
InternVL2\_5-8B & 79.3 & 29.1 & 36.1 & 27.9 & 32.1 & 67.9 \\
InterVL2-26B & 	\underline{81.7}	& 36.7	  &  58.9	 & 38.3 & 	\underline{58.1}	& 73.5 \\
LLaVA-32B & 76.4 & \underline{45.1} & \underline{81.2} & \underline{50.4} & 48.9 & \underline{87.4}  \\
\midrule
\multicolumn{7}{l}{\textbf{SFT Results}} \\
\midrule
Qwen2.5-VL-3B & 73.5 & 30.8 & 75.2 & 52.3 & 50.2 & 77.2 \\
Qwen2.5-VL-7B & 80.9 & 59.9 & 89.7 & 41.2 & 68.9 & 92.2 \\
InternVL2\_5-4B & 72.1 & 31.8 & 64.1 & 56.8 & 30.5 & 60.3 \\
\midrule
\multicolumn{7}{l}{\textbf{Ours}} \\
\midrule
\rowcolor{gray!10} Curr-ReFT-3B & {82.3} & {73.7} & {89.8} & {65.6} & {71.5} & {95.2} \\
\rowcolor{gray!10} Curr-ReFT-7B & \textbf{85.3} & \textbf{81.5} & \textbf{92.2} & \textbf{69.5} & \textbf{73.1} & \textbf{98.7} \\
\bottomrule
\end{tabular}
\caption{\textbf{Performance comparison on visual tasks. `In' denotes in-domain testing while `out' represents out-of-domain testing. The best results are in boldface and the second best results are underlined.}}
\label{tab:compare_methods}
\end{table}

First of all, we need to prove the effectiveness and generalization of RL methods, especially comparing with SFT methods. As a result, we train the models on math, classify, and detection tasks (specifically, refcoco, recocop, openr1\-8k datasets), with both SFT and GRPO methods. Then we evaluate the trained models on in-distribution and out-distribution datasets, respectively. As shown in Tab. \ref{tab:comprehensive}, we compare our trained Qwen2.5VL-3B with its variants of SFT, RL, and Curriculum Reinforcement Learning (Curr-RL). Moreover, we display qualitative examples between SFT and our Curr-Rl methods in Fig. \ref{compare_viosn}. Both quantitative and qualitative results indicate the following observations: 

\begin{itemize}
    \item In in-domain scenarios, RL-based methods clearly get more comvincing results than SFT, which shows the effectiveness and necessarity of incorporating RL-based methods into post-training stage.
    \item In out-of-domain scenarios, SFT shows limited improvements and even gets worse in some cases, while RL-based methods present strong generalization power, especifically with our introduced Curr-RL, which convinces us to apply Curr-RL to revolute the CV tasks for VLMs. 
    \item Our Curr-RL approach not only provides more detailed and comprehensive explanations but also achieves more accurate localization performance across out-of-domain visual tasks.
\end{itemize}

Besides, we give a line chart to observe the variation during RL training and SFT training. The results are reported in Fig. \ref{fig_sft_vs_grpo}, from which we have the following observations: 
\begin{itemize}
    \item With the training step increasing, both SFT and RL based methods present a satisfying improvement in in-domain evaluation. The SFT-based methods present a decreasing trend out-of-domain case, wihle the RL-based method still have a convining performance, which further presents the generalization ability.
    \item Overall, RL-based methods present the best results in terms of both in-domain and out-of-domain scenarios.
\end{itemize}

\subsection{Performance Comparation (RQ2)}

\begin{table*}[th]
\setlength{\tabcolsep}{6pt}  
\centering
\begin{tabular}{>{\raggedright\arraybackslash}p{2.6cm} >{\centering\arraybackslash}p{0.7cm} >{\centering\arraybackslash}p{0.9cm} >{\centering\arraybackslash}p{1.2cm} >{\centering\arraybackslash}p{1.1cm} >{\centering\arraybackslash}p{1.1cm} *{3}{>{\centering\arraybackslash}p{1.1cm}} >{\centering\arraybackslash}p{0.8cm}}
\toprule
\multirow{2}{*}{\textbf{Model}} & 
\multirow{2}{*}{\textbf{\small AI2D}} & 
\multirow{2}{*}{\textbf{\small MMVet}} & 
\multirow{2}{*}{\textbf{\small MathVista}} & 
\multirow{2}{*}{\textbf{\footnotesize OCRBench}} & 
\multirow{2}{*}{\textbf{\footnotesize MMBench}} & 
\multicolumn{3}{c}{\textbf{LLaVABench}} & 
\multirow{2}{*}{\textbf{\small MATH}} \\
\cmidrule(lr){7-9}
& & & & & & \textbf{\small Relative} & \textbf{\small VLM} & \textbf{\small GPT4} & \\
\midrule
Qwen2.5-VL-3B & 0.74 & 25.00 & 51.90 & 60.05 & 0.58 & 67.30 & 56.30 & 80.40 & 28.60 \\
InternVL2\_5-4B & 0.69 & 21.70 & 42.80 & 57.10 & 0.53 & 59.50 & 54.10 & 78.20 & 29.10 \\
Qwen2-VL-7B & 0.66 & 22.30 & 44.10 & 53.60 & 0.49 & 60.20 & 51.20 & 74.50 & 20.70 \\
Qwen2.5-VL-7B & 0.81 & 30.11 & \underline{60.70} & 64.12 & 0.67 & \underline{81.20} & 67.10 & 81.40 & 32.40 \\
InternVL2\_5-8B & 0.79 & 31.20 & 56.20 & 59.70 & 0.64 & 72.10 & 63.70 & 80.10 & 33.10 \\
InterVL2-26B & 0.78	&\textbf{36.10} &	56.4& \underline{70.10}&	0.68&	78.1& 74.11&81.00&	27.10 \\
LLaVA-32b &0.75	&30.20	&57.10	&63.2	&\underline{0.76} &	80.21 & \textbf{75.30}& 82.20 &	29.10\\
\rowcolor{gray!10}Curr-ReFT-3B  & \textbf{0.83} & 29.95 & 58.60 & 64.70 & 0.70 & 73.80 & 68.10 & \textbf{85.60} & \underline{35.80} \\
\rowcolor{gray!10}Curr-ReFT-7B  & \underline{0.82} & \underline{35.56} & \textbf{64.5} & \textbf{72.2}	& \textbf{0.79} & \textbf{83.60} & \underline{69.70} & \underline{84.50} & \textbf{45.60} \\
\bottomrule
\end{tabular}
\vspace{-5pt}
\caption{\textbf{Performance comparison of our approach against baseline models across multiple benchmarks. The best results are in boldface and the second best results are underlined.}}
\label{tab:compare_benchmark}
\end{table*}


We report the empirical results of all methods in Tab. \ref{tab:compare_methods} and Tab. \ref{tab:compare_benchmark}. Analyzing such performance comparison, we observate that:
\begin{itemize}
    \item \textbf{Models trained with our Curr-ReFT consistently demonstrate exceptional performance.} When comparing Curr-ReFT-3B with baseline models, our model performs extremely well across both our in-domain and out-of-domain datasets as well as on publicly recognized benchmarks. Remarkably, our 3B model frequently outperforms large 26B (InternVL-26B) and 32B (Llava-Next-32B) SOTA models, further demonstrating the effectiveness of our Curr-ReFT post-training framework.
    \item \textbf{Our trained models show significant improvements in Math and Logic capabilities.} When comparing Curr-ReFT-3B with Qwen2.5VL-3B, it is evident that our model achieves higher scores on math and reasoning benchmarks, such as AI2D and MMVet.
    \item \textbf{The generalization of our models has seen the most remarkable improvement.} Through the analysis of out-of-domain datasets and LLaVABench, we conclude that Curr-ReFT effectively enhances the generalization power of models, enabling the trained models to perform more satisfactorily than the 26B and 32B models.
     
\end{itemize}

\subsection{Ablation Study  (RQ3)}

\begin{table}[thbp]
\vspace{-10pt}
\setlength{\tabcolsep}{6pt}
\centering
\begin{subtable}{\linewidth}
\centering
\resizebox{\linewidth}{!}{
\begin{tabular}{@{}ccccccc@{}}
\toprule
\multirow{2}{*}{\textbf{Method}} & \multicolumn{2}{c}{\textbf{Math}} & \multicolumn{2}{c}{\textbf{Detection}} & \multicolumn{2}{c}{\textbf{Classification}} \\
\cmidrule(lr){2-3} \cmidrule(lr){4-5} \cmidrule(lr){6-7}
& \textbf{In} & \textbf{Out} & \textbf{In} & \textbf{Out} & \textbf{In} & \textbf{Out} \\ 
\midrule
\textbf{Base} & 71.3 & 17.8 & 31.8 & 22.3 & 39.6 & 79.8 \\[0.5ex]
\textbf{+SFT} & 73.5 & 30.8 & 75.2 & 52.3 & 50.2 & 77.2 \\[0.5ex]
\textbf{+RL} & 78.8 & 74.1 & 88.3 & 64.2 & 62.9 & 94.7 \\[0.5ex]
\textbf{+Curr-RL} & \textbf{82.8} & \textbf{78.4} & \textbf{90.6} & \textbf{67.1} & \textbf{66.8} & \textbf{96.6} \\[0.5ex]
\textbf{+Curr-ReFT} & 82.3 & 73.7 & 89.8 & 65.6 & 65.4 & 95.2 \\
\bottomrule
\end{tabular}
}
\caption{\textbf{Performance on visual perception and math datasets.}}
\label{tab:ablation_visual}
\end{subtable}

\vspace{1em}

\begin{subtable}{\linewidth}
\centering
\resizebox{1.00\linewidth}{!}
{  
\begin{tabular}{@{}cccccccc@{}}
\toprule
\multirow{3}{*}{\textbf{\Large Method}} & 
\multirow{3}{*}{\textbf{\Large AI2D}} & 
\multirow{3}{*}{\textbf{\Large MMVet}} & 
\multirow{3}{*}{\textbf{\Large MathV}} & 
\multirow{3}{*}{\textbf{\Large OCR}} & 
\multirow{3}{*}{\textbf{\Large MM}} & 
\multicolumn{2}{c}{\textbf{\Large LLaVA}} \\[0.8ex]  
\cmidrule(lr){7-8}
& & & & & & \textbf{\Large VLM} & \textbf{\Large GPT4} \\[0.8ex]
\midrule
\textbf{\Large Base} & {\Large 0.71} & {\Large 25.00} & {\Large 51.90} & {\Large 60.05} & {\Large 0.58} & {\Large 56.30} & {\Large 80.40} \\[1.5ex]
\textbf{\Large +SFT} & {\Large 0.66} & {\Large 25.50} & {\Large 49.50} & {\Large 56.50} & {\Large 0.57} & {\Large 52.30} & {\Large 84.10} \\[1.5ex]
\textbf{\Large +RL} & {\Large 0.76} & {\Large 26.24} & {\Large 52.50} & {\Large 59.50} & {\Large 0.65} & {\Large 57.70} & {\Large 83.00} \\[1.5ex]
\textbf{\Large +Curr-RL} & {\Large 0.78} & {\Large 27.43} & {\Large 56.90} & {\Large 60.80} & {\Large 0.67} & {\Large 57.10} & {\Large \textbf{85.20}} \\[1.5ex]
\textbf{\Large +Curr-ReFT} & {\Large \textbf{0.83}} & {\Large \textbf{29.95}} & {\Large \textbf{58.60}} & {\Large \textbf{63.70}} & {\Large \textbf{0.70}} & {\Large \textbf{68.10}} & {\Large 83.60} \\
\bottomrule
\end{tabular}
}
\caption{\textbf{Performance on standard benchmarks}}
\end{subtable}
\caption{\textbf{Ablation Study on both CV tasks and benchmarks.}}
\label{tab:ablation_full}
\end{table}

As shown in Tab. \ref{tab:ablation_full} (a) and Tab. \ref{tab:ablation_full} (b), here we examine the contributions of main components in Curr-ReFT, by comparing the Qwen2.5-VL-3B+Curr-ReFT with the following two variants: 1) $\text{Qwen2.5-VL-3B+Curr-RL}$: In this variant, the RST is removed. 2) $\text{Qwen2.5-VL-3B+RL}$: This variant removes Curriculum Reinforcement Learning, from which we have the following observations: 
\begin{itemize}
\item Removing both Reinforcement Learning and Reject-sample SFT would degrade model performance, demonstrating their effectiveness in the post-training of VLMs.
\item Ablating the Curriculum RL leads to the worst performance, which underscores the critical role of incorporating a well-structured reinforcement learning component.
\item The integration of Rejected Sample based Self-improvement brings a trade-off. It slightly reduces task-specific performance (like detection and classification), but greatly enhances the general capabilities, especially in reasoning and generalization benchmarks. This shows Curr-ReFT effectively re-balances the model, trading minimal task-specific performance loss for improved general capabilities.
     
\end{itemize}

\subsection{Scaling Analysis (RQ4)}

To examine the scaling effectiveness of our Curr-ReFT framework, we carried out extensive experiments on the larger Qwen2.5-VL-7B model. The results in Tab. \ref{tab:compare_methods} and Tab. \ref{tab:compare_benchmark} indicate that the effectiveness of Curr-ReFT scales effectively with model size:
\begin{itemize}
\item Compared to the Curr-ReFT-3B, Curr-ReFT-7B shows consistent improvements: Higher performance on visual tasks (detection: 89.8\% → 92.2\%, classification: 71.5\% → 73.1\%) and better generalization on benchmarks (MMVet: 29.95\% → 36.78\%, MathVista: 58.60\% → 92.2\%).
\item The performance gains are particularly pronounced in complex reasoning tasks, suggesting that larger models better leverage our Curr-ReFT approach for reasoning.
\end{itemize}

\section{Conclusion}\label{sec:conclusion}

In this paper, we focus on improving both reasoning and OOD generalization capabilities of small-scale VLMs. Our empirical findings reveal that reinforcement learning not only enhances reasoning abilities but also surprisingly improves generalization in visual tasks. Based on these insights, we propose Curriculum Reinforcement Finetuning (Curr-ReFT), a novel post-training paradigm that combines progressive curriculum learning with rejected sampling. Through gradually increasing task complexity and selective learning from high-quality examples, Curr-ReFT enables stable optimization while maintaining both reasoning and generalization capabilities.
{
    \small
    \bibliographystyle{ieeenat_fullname}
    \bibliography{main}

\begin{thebibliography}{49}
\providecommand{\natexlab}[1]{#1}
\providecommand{\url}[1]{\texttt{#1}}
\expandafter\ifx\csname urlstyle\endcsname\relax
  \providecommand{\doi}[1]{doi: #1}\else
  \providecommand{\doi}{doi: \begingroup \urlstyle{rm}\Url}\fi

\bibitem[Arjovsky(2020)]{arjovsky2020ood}
Martin Arjovsky.
\newblock \emph{Out of distribution generalization in machine learning}.
\newblock PhD thesis, New York University, 2020.

\bibitem[Arrieta et~al.(2025)Arrieta, Ugarte, Valle, Parejo, and Segura]{openai_o3}
Aitor Arrieta, Miriam Ugarte, Pablo Valle, Jos{\'e}~Antonio Parejo, and Sergio Segura.
\newblock Early external safety testing of openai's o3-mini: Insights from the pre-deployment evaluation.
\newblock \emph{arXiv preprint arXiv:2501.17749}, 2025.

\bibitem[Bai et~al.(2023)Bai, Bai, Chu, Cui, Dang, Deng, Fan, Ge, Han, Huang, et~al.]{bai2023qwen}
Jinze Bai, Shuai Bai, Yunfei Chu, Zeyu Cui, Kai Dang, Xiaodong Deng, Yang Fan, Wenbin Ge, Yu Han, Fei Huang, et~al.
\newblock Qwen technical report.
\newblock \emph{arXiv preprint arXiv:2309.16609}, 2023.

\bibitem[Bai et~al.(2022)Bai, Kadavath, Kundu, Askell, Kernion, Jones, Chen, Goldie, Mirhoseini, McKinnon, et~al.]{bai2022sft}
Yuntao Bai, Saurav Kadavath, Sandipan Kundu, Amanda Askell, Jackson Kernion, Andy Jones, Anna Chen, Anna Goldie, Azalia Mirhoseini, Cameron McKinnon, et~al.
\newblock Constitutional ai: Harmlessness from ai feedback.
\newblock \emph{arXiv preprint arXiv:2212.08073}, 2022.

\bibitem[Brown et~al.(2020)]{brown2020language}
Tom Brown et~al.
\newblock Language models are few-shot learners.
\newblock \emph{Advances in neural information processing systems}, 2020.

\bibitem[Browne et~al.(2012)Browne, Powley, Whitehouse, Lucas, Cowling, Rohlfshagen, Tavener, Perez, Samothrakis, and Colton]{browne2012survey}
Cameron~B Browne, Edward Powley, Daniel Whitehouse, Simon~M Lucas, Peter~I Cowling, Philipp Rohlfshagen, Stephen Tavener, Diego Perez, Spyridon Samothrakis, and Simon Colton.
\newblock A survey of monte carlo tree search methods.
\newblock \emph{IEEE Transactions on Computational Intelligence and AI in games}, 4\penalty0 (1):\penalty0 1--43, 2012.

\bibitem[Casper et~al.(2023)Casper, Davies, Shi, Gilbert, Scheurer, Rando, Freedman, Korbak, Lindner, Freire, et~al.]{casper2023open}
Stephen Casper, Xander Davies, Claudia Shi, Thomas~Krendl Gilbert, J{\'e}r{\'e}my Scheurer, Javier Rando, Rachel Freedman, Tomasz Korbak, David Lindner, Pedro Freire, et~al.
\newblock Open problems and fundamental limitations of reinforcement learning from human feedback.
\newblock \emph{arXiv preprint arXiv:2307.15217}, 2023.

\bibitem[Chen et~al.(2024{\natexlab{a}})Chen, Wang, Cao, Liu, Gao, Cui, Zhu, Ye, Tian, Liu, et~al.]{chen2024expanding}
Zhe Chen, Weiyun Wang, Yue Cao, Yangzhou Liu, Zhangwei Gao, Erfei Cui, Jinguo Zhu, Shenglong Ye, Hao Tian, Zhaoyang Liu, et~al.
\newblock Expanding performance boundaries of open-source multimodal models with model, data, and test-time scaling.
\newblock \emph{arXiv preprint arXiv:2412.05271}, 2024{\natexlab{a}}.

\bibitem[Chen et~al.(2024{\natexlab{b}})Chen, Wu, Wang, Su, Chen, Xing, Zhong, Zhang, Zhu, Lu, et~al.]{chen2024internvl}
Zhe Chen, Jiannan Wu, Wenhai Wang, Weijie Su, Guo Chen, Sen Xing, Muyan Zhong, Qinglong Zhang, Xizhou Zhu, Lewei Lu, et~al.
\newblock Internvl: Scaling up vision foundation models and aligning for generic visual-linguistic tasks.
\newblock In \emph{Proceedings of the IEEE/CVF conference on computer vision and pattern recognition}, pages 24185--24198, 2024{\natexlab{b}}.

\bibitem[Christiano et~al.(2017)]{christiano2017deep}
Paul~F Christiano et~al.
\newblock Deep reinforcement learning from human preferences.
\newblock \emph{Advances in neural information processing systems}, 2017.

\bibitem[Chung et~al.(2024)Chung, Hou, Longpre, Zoph, Tay, Fedus, Li, Wang, Dehghani, Brahma, et~al.]{instruction-finetuned}
Hyung~Won Chung, Le Hou, Shayne Longpre, Barret Zoph, Yi Tay, William Fedus, Yunxuan Li, Xuezhi Wang, Mostafa Dehghani, Siddhartha Brahma, et~al.
\newblock Scaling instruction-finetuned language models.
\newblock \emph{Journal of Machine Learning Research}, 25\penalty0 (70):\penalty0 1--53, 2024.

\bibitem[Everingham et~al.(2010)Everingham, Van~Gool, Williams, Winn, and Zisserman]{pascal_voc}
Mark Everingham, Luc Van~Gool, Christopher~KI Williams, John Winn, and Andrew Zisserman.
\newblock The pascal visual object classes (voc) challenge.
\newblock \emph{International journal of computer vision}, 88:\penalty0 303--338, 2010.

\bibitem[Gao et~al.(2023)Gao, Pi, Zhang, Ye, Zhong, Wang, Hong, Han, Xu, Li, et~al.]{gao2023geo170k}
Jiahui Gao, Renjie Pi, Jipeng Zhang, Jiacheng Ye, Wanjun Zhong, Yufei Wang, Lanqing Hong, Jianhua Han, Hang Xu, Zhenguo Li, et~al.
\newblock G-llava: Solving geometric problem with multi-modal large language model.
\newblock \emph{arXiv preprint arXiv:2312.11370}, 2023.

\bibitem[Gao et~al.(2025)Gao, Jin, Ke, and Moryoussef]{gao2025comparison}
Tianchen Gao, Jiashun Jin, Zheng~Tracy Ke, and Gabriel Moryoussef.
\newblock A comparison of deepseek and other llms.
\newblock \emph{arXiv preprint arXiv:2502.03688}, 2025.

\bibitem[Guo et~al.(2025)Guo, Yang, Zhang, Song, Zhang, Xu, Zhu, Ma, Wang, Bi, et~al.]{guo2025deepseek}
Daya Guo, Dejian Yang, Haowei Zhang, Junxiao Song, Ruoyu Zhang, Runxin Xu, Qihao Zhu, Shirong Ma, Peiyi Wang, Xiao Bi, et~al.
\newblock Deepseek-r1: Incentivizing reasoning capability in llms via reinforcement learning.
\newblock \emph{arXiv preprint arXiv:2501.12948}, 2025.

\bibitem[Hendrycks et~al.(2021)Hendrycks, Burns, Kadavath, Arora, Basart, Tang, Song, and Steinhardt]{MATH}
Dan Hendrycks, Collin Burns, Saurav Kadavath, Akul Arora, Steven Basart, Eric Tang, Dawn Song, and Jacob Steinhardt.
\newblock Measuring mathematical problem solving with the math dataset.
\newblock \emph{arXiv preprint arXiv:2103.03874}, 2021.

\bibitem[Hiippala et~al.(2021)Hiippala, Alikhani, Haverinen, Kalliokoski, Logacheva, Orekhova, Tuomainen, Stone, and Bateman]{hiippala2021ai2d}
Tuomo Hiippala, Malihe Alikhani, Jonas Haverinen, Timo Kalliokoski, Evanfiya Logacheva, Serafina Orekhova, Aino Tuomainen, Matthew Stone, and John~A Bateman.
\newblock Ai2d-rst: a multimodal corpus of 1000 primary school science diagrams.
\newblock \emph{Language Resources and Evaluation}, 55:\penalty0 661--688, 2021.

\bibitem[Huang et~al.(2020)Huang, Wang, Tai, Liu, Shen, Li, Li, and Huang]{huang2020curricularface}
Yuge Huang, Yuhan Wang, Ying Tai, Xiaoming Liu, Pengcheng Shen, Shaoxin Li, Jilin Li, and Feiyue Huang.
\newblock Curricularface: adaptive curriculum learning loss for deep face recognition.
\newblock In \emph{proceedings of the IEEE/CVF conference on computer vision and pattern recognition}, pages 5901--5910, 2020.

\bibitem[Jaech et~al.(2024)Jaech, Kalai, Lerer, Richardson, El-Kishky, Low, Helyar, Madry, Beutel, Carney, et~al.]{openai_O1}
Aaron Jaech, Adam Kalai, Adam Lerer, Adam Richardson, Ahmed El-Kishky, Aiden Low, Alec Helyar, Aleksander Madry, Alex Beutel, Alex Carney, et~al.
\newblock Openai o1 system card.
\newblock \emph{arXiv preprint arXiv:2412.16720}, 2024.

\bibitem[Jiang et~al.(2015)Jiang, Meng, Zhao, Shan, and Hauptmann]{jiang2015self}
Lu Jiang, Deyu Meng, Qian Zhao, Shiguang Shan, and Alexander Hauptmann.
\newblock Self-paced curriculum learning.
\newblock In \emph{Proceedings of the AAAI Conference on Artificial Intelligence}, 2015.

\bibitem[Kong et~al.(2021)Kong, Liu, Wang, and Tao]{kong2021adaptive}
Yajing Kong, Liu Liu, Jun Wang, and Dacheng Tao.
\newblock Adaptive curriculum learning.
\newblock In \emph{Proceedings of the IEEE/CVF International Conference on Computer Vision}, pages 5067--5076, 2021.

\bibitem[Liu et~al.(2024{\natexlab{a}})Liu, Feng, Wang, Wang, Liu, Zhao, Dengr, Ruan, Dai, Guo, et~al.]{liu2024deepseekv2}
Aixin Liu, Bei Feng, Bin Wang, Bingxuan Wang, Bo Liu, Chenggang Zhao, Chengqi Dengr, Chong Ruan, Damai Dai, Daya Guo, et~al.
\newblock Deepseek-v2: A strong, economical, and efficient mixture-of-experts language model.
\newblock \emph{arXiv preprint arXiv:2405.04434}, 2024{\natexlab{a}}.

\bibitem[Liu et~al.(2024{\natexlab{b}})Liu, Feng, Xue, Wang, Wu, Lu, Zhao, Deng, Zhang, Ruan, et~al.]{liu2024deepseekv3}
Aixin Liu, Bei Feng, Bing Xue, Bingxuan Wang, Bochao Wu, Chengda Lu, Chenggang Zhao, Chengqi Deng, Chenyu Zhang, Chong Ruan, et~al.
\newblock Deepseek-v3 technical report.
\newblock \emph{arXiv preprint arXiv:2412.19437}, 2024{\natexlab{b}}.

\bibitem[Liu et~al.(2023)Liu, Li, Wu, and Lee]{llava}
Haotian Liu, Chunyuan Li, Qingyang Wu, and Yong~Jae Lee.
\newblock Visual instruction tuning.
\newblock \emph{Advances in neural information processing systems}, 36:\penalty0 34892--34916, 2023.

\bibitem[Liu et~al.(2021)Liu, Shen, He, Zhang, Xu, Yu, and Cui]{liu2021towards}
Jiashuo Liu, Zheyan Shen, Yue He, Xingxuan Zhang, Renzhe Xu, Han Yu, and Peng Cui.
\newblock Towards out-of-distribution generalization: A survey.
\newblock \emph{arXiv preprint arXiv:2108.13624}, 2021.

\bibitem[Liu et~al.(2024{\natexlab{c}})Liu, Duan, Zhang, Li, Zhang, Zhao, Yuan, Wang, He, Liu, et~al.]{liu2024mmbench}
Yuan Liu, Haodong Duan, Yuanhan Zhang, Bo Li, Songyang Zhang, Wangbo Zhao, Yike Yuan, Jiaqi Wang, Conghui He, Ziwei Liu, et~al.
\newblock Mmbench: Is your multi-modal model an all-around player?
\newblock In \emph{European conference on computer vision}, pages 216--233. Springer, 2024{\natexlab{c}}.

\bibitem[Liu et~al.(2024{\natexlab{d}})Liu, Li, Huang, Yang, Yu, Li, Yin, Liu, Jin, and Bai]{liu2024ocrbench}
Yuliang Liu, Zhang Li, Mingxin Huang, Biao Yang, Wenwen Yu, Chunyuan Li, Xu-Cheng Yin, Cheng-Lin Liu, Lianwen Jin, and Xiang Bai.
\newblock Ocrbench: on the hidden mystery of ocr in large multimodal models.
\newblock \emph{Science China Information Sciences}, 67\penalty0 (12):\penalty0 220102, 2024{\natexlab{d}}.

\bibitem[Lu et~al.(2024)Lu, Dou, Wang, Cao, Dai, Feng, and Guo]{lu2024autopsv}
Jianqiao Lu, Zhiyang Dou, Hongru Wang, Zeyu Cao, Jianbo Dai, Yunlong Feng, and Zhijiang Guo.
\newblock Autopsv: Automated process-supervised verifier.
\newblock \emph{Advances in Neural Information Processing Systems}, 37:\penalty0 79935--79962, 2024.

\bibitem[Lu et~al.(2023)Lu, Bansal, Xia, Liu, Li, Hajishirzi, Cheng, Chang, Galley, and Gao]{lu2023mathvista}
Pan Lu, Hritik Bansal, Tony Xia, Jiacheng Liu, Chunyuan Li, Hannaneh Hajishirzi, Hao Cheng, Kai-Wei Chang, Michel Galley, and Jianfeng Gao.
\newblock Mathvista: Evaluating mathematical reasoning of foundation models in visual contexts.
\newblock \emph{arXiv preprint arXiv:2310.02255}, 2023.

\bibitem[Luo et~al.(2024)Luo, Yang, Dou, Wang, Dai, Qiao, and Zhu]{luo2024mono}
Gen Luo, Xue Yang, Wenhan Dou, Zhaokai Wang, Jifeng Dai, Yu Qiao, and Xizhou Zhu.
\newblock Mono-internvl: Pushing the boundaries of monolithic multimodal large language models with endogenous visual pre-training.
\newblock \emph{arXiv preprint arXiv:2410.08202}, 2024.

\bibitem[Mao et~al.(2016)Mao, Huang, Toshev, Camburu, Yuille, and Murphy]{Refcocog}
Junhua Mao, Jonathan Huang, Alexander Toshev, Oana Camburu, Alan~L Yuille, and Kevin Murphy.
\newblock Generation and comprehension of unambiguous object descriptions.
\newblock In \emph{Proceedings of the IEEE conference on computer vision and pattern recognition}, pages 11--20, 2016.

\bibitem[Ouyang et~al.(2022)Ouyang, Wu, Jiang, Almeida, Wainwright, Mishkin, Zhang, Agarwal, Slama, Ray, et~al.]{ouyang2022training}
Long Ouyang, Jeffrey Wu, Xu Jiang, Diogo Almeida, Carroll Wainwright, Pamela Mishkin, Chong Zhang, Sandhini Agarwal, Katarina Slama, Alex Ray, et~al.
\newblock Training language models to follow instructions with human feedback.
\newblock \emph{Advances in neural information processing systems}, 35:\penalty0 27730--27744, 2022.

\bibitem[Pentina et~al.(2015)Pentina, Sharmanska, and Lampert]{pentina2015curriculum}
Anastasia Pentina, Viktoriia Sharmanska, and Christoph~H Lampert.
\newblock Curriculum learning of multiple tasks.
\newblock In \emph{Proceedings of the IEEE conference on computer vision and pattern recognition}, pages 5492--5500, 2015.

\bibitem[Radford et~al.(2021)Radford, Kim, Hallacy, Ramesh, Goh, Agarwal, Sastry, Askell, Mishkin, Clark, et~al.]{clip}
Alec Radford, Jong~Wook Kim, Chris Hallacy, Aditya Ramesh, Gabriel Goh, Sandhini Agarwal, Girish Sastry, Amanda Askell, Pamela Mishkin, Jack Clark, et~al.
\newblock Learning transferable visual models from natural language supervision.
\newblock In \emph{International conference on machine learning}, pages 8748--8763. PmLR, 2021.

\bibitem[Schulman et~al.(2017)]{schulman2017proximal}
John Schulman et~al.
\newblock Proximal policy optimization algorithms.
\newblock \emph{arXiv preprint arXiv:1707.06347}, 2017.

\bibitem[Shi et~al.(2024)Shi, Hu, Bin, Liu, Yang, Ng, Bing, and Lee]{shi2024math360k}
Wenhao Shi, Zhiqiang Hu, Yi Bin, Junhua Liu, Yang Yang, See-Kiong Ng, Lidong Bing, and Roy Ka-Wei Lee.
\newblock Math-llava: Bootstrapping mathematical reasoning for multimodal large language models.
\newblock \emph{arXiv preprint arXiv:2406.17294}, 2024.

\bibitem[{\'S}wiechowski et~al.(2023){\'S}wiechowski, Godlewski, Sawicki, and Ma{\'n}dziuk]{swiechowski2023monte}
Maciej {\'S}wiechowski, Konrad Godlewski, Bartosz Sawicki, and Jacek Ma{\'n}dziuk.
\newblock Monte carlo tree search: A review of recent modifications and applications.
\newblock \emph{Artificial Intelligence Review}, 56\penalty0 (3):\penalty0 2497--2562, 2023.

\bibitem[Tanaka et~al.(2019)Tanaka, Itamochi, Narioka, Sato, Ushiku, and Harada]{Refgta}
Mikihiro Tanaka, Takayuki Itamochi, Kenichi Narioka, Ikuro Sato, Yoshitaka Ushiku, and Tatsuya Harada.
\newblock Generating easy-to-understand referring expressions for target identifications.
\newblock In \emph{Proceedings of the IEEE/CVF International Conference on Computer Vision}, pages 5794--5803, 2019.

\bibitem[Wainwright and Lowe(2023)]{wainwright2023instructgpt}
Carroll Wainwright and Ryan Lowe.
\newblock Instructgpt: Training language models to follow instructions with human feedback.
\newblock \emph{GitHub repository}, 2023.

\bibitem[Wang et~al.(2024)Wang, Bai, Tan, Wang, Fan, Bai, Chen, Liu, Wang, Ge, et~al.]{Qwen2VL}
Peng Wang, Shuai Bai, Sinan Tan, Shijie Wang, Zhihao Fan, Jinze Bai, Keqin Chen, Xuejing Liu, Jialin Wang, Wenbin Ge, et~al.
\newblock Qwen2-vl: Enhancing vision-language model's perception of the world at any resolution.
\newblock \emph{arXiv preprint arXiv:2409.12191}, 2024.

\bibitem[Wang et~al.(2023)Wang, Bao, Dong, Bjorck, Peng, Liu, Aggarwal, Mohammed, Singhal, Som, et~al.]{Beit}
Wenhui Wang, Hangbo Bao, Li Dong, Johan Bjorck, Zhiliang Peng, Qiang Liu, Kriti Aggarwal, Owais~Khan Mohammed, Saksham Singhal, Subhojit Som, et~al.
\newblock Image as a foreign language: Beit pretraining for vision and vision-language tasks.
\newblock In \emph{Proceedings of the IEEE/CVF Conference on Computer Vision and Pattern Recognition}, pages 19175--19186, 2023.

\bibitem[Wang et~al.(2022{\natexlab{a}})Wang, Wei, Schuurmans, Le, Chi, Narang, Chowdhery, and Zhou]{wang2022self_cot}
Xuezhi Wang, Jason Wei, Dale Schuurmans, Quoc Le, Ed Chi, Sharan Narang, Aakanksha Chowdhery, and Denny Zhou.
\newblock Self-consistency improves chain of thought reasoning in language models.
\newblock \emph{arXiv preprint arXiv:2203.11171}, 2022{\natexlab{a}}.

\bibitem[Wang et~al.(2022{\natexlab{b}})Wang, Kordi, Mishra, Liu, Smith, Khashabi, and Hajishirzi]{Self-instruct}
Yizhong Wang, Yeganeh Kordi, Swaroop Mishra, Alisa Liu, Noah~A Smith, Daniel Khashabi, and Hannaneh Hajishirzi.
\newblock Self-instruct: Aligning language models with self-generated instructions.
\newblock \emph{arXiv preprint arXiv:2212.10560}, 2022{\natexlab{b}}.

\bibitem[Wei et~al.(2022)Wei, Wang, Schuurmans, Bosma, Xia, Chi, Le, Zhou, et~al.]{cot}
Jason Wei, Xuezhi Wang, Dale Schuurmans, Maarten Bosma, Fei Xia, Ed Chi, Quoc~V Le, Denny Zhou, et~al.
\newblock Chain-of-thought prompting elicits reasoning in large language models.
\newblock \emph{Advances in neural information processing systems}, 35:\penalty0 24824--24837, 2022.

\bibitem[Yang et~al.(2024)Yang, Yang, Zhang, Hui, Zheng, Yu, Li, Liu, Huang, Wei, et~al.]{qwen2_5}
An Yang, Baosong Yang, Beichen Zhang, Binyuan Hui, Bo Zheng, Bowen Yu, Chengyuan Li, Dayiheng Liu, Fei Huang, Haoran Wei, et~al.
\newblock Qwen2. 5 technical report.
\newblock \emph{arXiv preprint arXiv:2412.15115}, 2024.

\bibitem[Yao et~al.(2023)Yao, Yu, Zhao, Shafran, Griffiths, Cao, and Narasimhan]{ToT}
Shunyu Yao, Dian Yu, Jeffrey Zhao, Izhak Shafran, Tom Griffiths, Yuan Cao, and Karthik Narasimhan.
\newblock Tree of thoughts: Deliberate problem solving with large language models.
\newblock \emph{Advances in neural information processing systems}, 36:\penalty0 11809--11822, 2023.

\bibitem[Yu et~al.(2016)Yu, Poirson, Yang, Berg, and Berg]{Refcoco}
Licheng Yu, Patrick Poirson, Shan Yang, Alexander~C Berg, and Tamara~L Berg.
\newblock Modeling context in referring expressions.
\newblock In \emph{Computer Vision--ECCV 2016: 14th European Conference, Amsterdam, The Netherlands, October 11-14, 2016, Proceedings, Part II 14}, pages 69--85. Springer, 2016.

\bibitem[Yu et~al.(2023)Yu, Yang, Li, Wang, Lin, Liu, Wang, and Wang]{yu2023mm}
Weihao Yu, Zhengyuan Yang, Linjie Li, Jianfeng Wang, Kevin Lin, Zicheng Liu, Xinchao Wang, and Lijuan Wang.
\newblock Mm-vet: Evaluating large multimodal models for integrated capabilities.
\newblock \emph{arXiv preprint arXiv:2308.02490}, 2023.

\bibitem[Ziegler et~al.(2019)Ziegler, Stiennon, Wu, Brown, Radford, Amodei, Christiano, and Irving]{ziegler2019fine}
Daniel~M Ziegler, Nisan Stiennon, Jeffrey Wu, Tom~B Brown, Alec Radford, Dario Amodei, Paul Christiano, and Geoffrey Irving.
\newblock Fine-tuning language models from human preferences.
\newblock \emph{arXiv preprint arXiv:1909.08593}, 2019.

\end{thebibliography}
}

\end{document}